\documentclass[11pt,a4paper]{article}
\usepackage[hyperref]{emnlp2020}
\usepackage{times}
\usepackage{latexsym}

\usepackage{graphicx}
\usepackage{subcaption}
\usepackage{times}
\usepackage{url}
\usepackage{latexsym}
\usepackage{graphicx}
\usepackage{booktabs}
\usepackage{multirow}
\usepackage{mathrsfs}
\usepackage{amsmath}
\usepackage{bbm}
\usepackage{amssymb}
\usepackage{wrapfig}
\usepackage{balance}
\usepackage{url}
\usepackage{bm}
\usepackage{float}
\usepackage{amsmath}
\usepackage{amsthm}
\usepackage{amsfonts}
\usepackage{graphicx}
\usepackage{booktabs}
\usepackage{placeins}
\usepackage{multirow}
\usepackage{epstopdf}
\usepackage{color}
\usepackage[T1]{fontenc}
\usepackage{balance}
\usepackage{enumitem}
\usepackage{xcolor}
\usepackage{balance}
\usepackage{relsize}
\usepackage{enumitem}
\usepackage{wrapfig}
\usepackage{graphics}
\usepackage{multirow}
\usepackage{url}
\usepackage{makecell}
\usepackage{array}
\newcommand{\PreserveBackslash}[1]{\let\temp=\\#1\let\\=\temp}
\newcolumntype{C}[1]{>{\PreserveBackslash\centering}p{#1}}
\newcolumntype{R}[1]{>{\PreserveBackslash\raggedleft}p{#1}}
\newcolumntype{L}[1]{>{\PreserveBackslash\raggedright}p{#1}}

\definecolor{darkblue}{rgb}{0, 0, 0.5}
\hypersetup{colorlinks=true,citecolor=darkblue, linkcolor=darkblue, urlcolor=darkblue}

\aclfinalcopy 

\title{
Using Type Information to Improve Entity Coreference Resolution
}

\author{Sopan Khosla \quad \quad \textbf{Carolyn Rose}\\
	\vspace{-4 mm} \\
	Language Technologies Institute \\
	Carnegie Mellon University, USA \\
	{\tt \small \{sopank, cprose\}@cs.cmu.edu} \\
}
\date{}

\begin{document}
	\newcommand{\refalg}[1]{Algorithm \ref{#1}}
\newcommand{\refeqn}[1]{Equation \ref{#1}}
\newcommand{\reffig}[1]{Figure \ref{#1}}
\newcommand{\reftbl}[1]{Table \ref{#1}}
\newcommand{\refsec}[1]{Section \ref{#1}}

\newcommand{\add}[1]{\textcolor{red}{#1}\typeout{#1}}
\newcommand{\remove}[1]{\sout{#1}\typeout{#1}}

\newcommand{\m}[1]{\mathcal{#1}}
\newcommand{\bmm}[1]{\bm{\mathcal{#1}}}
\newcommand{\real}[1]{\mathbb{R}^{#1}}
\newcommand{\method}{\textsc{MedFilter}}

\newcommand{\bleuone}{BLEU${_1}$\xspace}
\newcommand{\bleu}{BLEU${_4}$\xspace}

\newcommand{\problem}{}
\newcommand{\problemfull}{}

\newtheorem{theorem}{Theorem}[section]
\newtheorem{claim}[theorem]{Claim}

\newcommand{\reminder}[1]{\textcolor{red}{[[ #1 ]]}\typeout{#1}}
\newcommand{\reminderR}[1]{\textcolor{gray}{[[ #1 ]]}\typeout{#1}}

\newcommand{\tensor}{\mathcal{X}}
\newcommand{\Real}{\mathbb{R}}

\newcommand{\tuples}{\mathbb{T}}

\newcommand\norm[1]{\left\lVert#1\right\rVert}

\newcommand{\note}[1]{\textcolor{blue}{#1}}

\newcommand*{\Scale}[2][4]{\scalebox{#1}{$#2$}}%
\newcommand*{\Resize}[2]{\resizebox{#1}{!}{$#2$}}%

\def\mat#1{\mbox{\bf #1}}
\newcommand{\cev}[1]{\reflectbox{\ensuremath{\vec{\reflectbox{\ensuremath{#1}}}}}}

	\maketitle
	\begin{abstract}
		Coreference resolution (CR) is an essential part of discourse analysis. Most recently, neural approaches have been proposed to improve over SOTA models from earlier paradigms. So far none of the published neural models leverage external semantic knowledge such as type information.  This paper offers the first such model and evaluation, demonstrating modest gains in accuracy by introducing either gold standard or predicted types.  In the proposed approach, type information serves both to (1) improve mention representation and (2) create a soft type consistency check between coreference candidate mentions. Our evaluation covers two different grain sizes of types over four different benchmark corpora.  
	\end{abstract}

	\section{Introduction}
\label{sec:intro}

Coreference resolution (CR) is an extensively studied problem in computational linguistics and NLP~\cite{coref_cl1, coref_cl2, coref_cl3,  coref_cl5, coref_cl4, lee2017end}. Solutions to this problem allow us to make meaningful links between concepts and entities within a discourse and therefore serves as a valuable pre-processing step for downstream tasks like summarization and question-answering~\cite{coref_summ1,Dasigi2019Quoref,coref_survey}.

Recently, multiple datasets including Ontonotes~\cite{onto2012}, Litbank~\cite{bamman2020annotated}, EmailCoref~\cite{dakle2020study}, and WikiCoref~\cite{ghaddar2016wikicoref} have been proposed as benchmark datasets for CR, especially in the sub-area of entity anaphora~\cite{sukthanker2020anaphora}.  Entity anaphora is a simpler starting place for work on anaphora because unlike abstract anaphora~\cite{webber1991abstract}, entity anaphora are pronouns or noun phrases that refer to an explicitly mentioned entity in the discourse rather than an abstract idea that must be constructed from a repackaging of information revealed over an extended text. An affordance of entity anaphora is that they have easily articulated semantic types.  Most of the entity CR datasets are extensively annotated for syntactic features (like constituency parse etc.) and semantic features (like entity-types). However, none of the published SOTA methods~\cite{lee2017end, joshi2019bert, joshi2020spanbert} explicitly leverage the type information.

\begin{figure*}
    \centering
    \includegraphics[width = 0.74\textwidth]{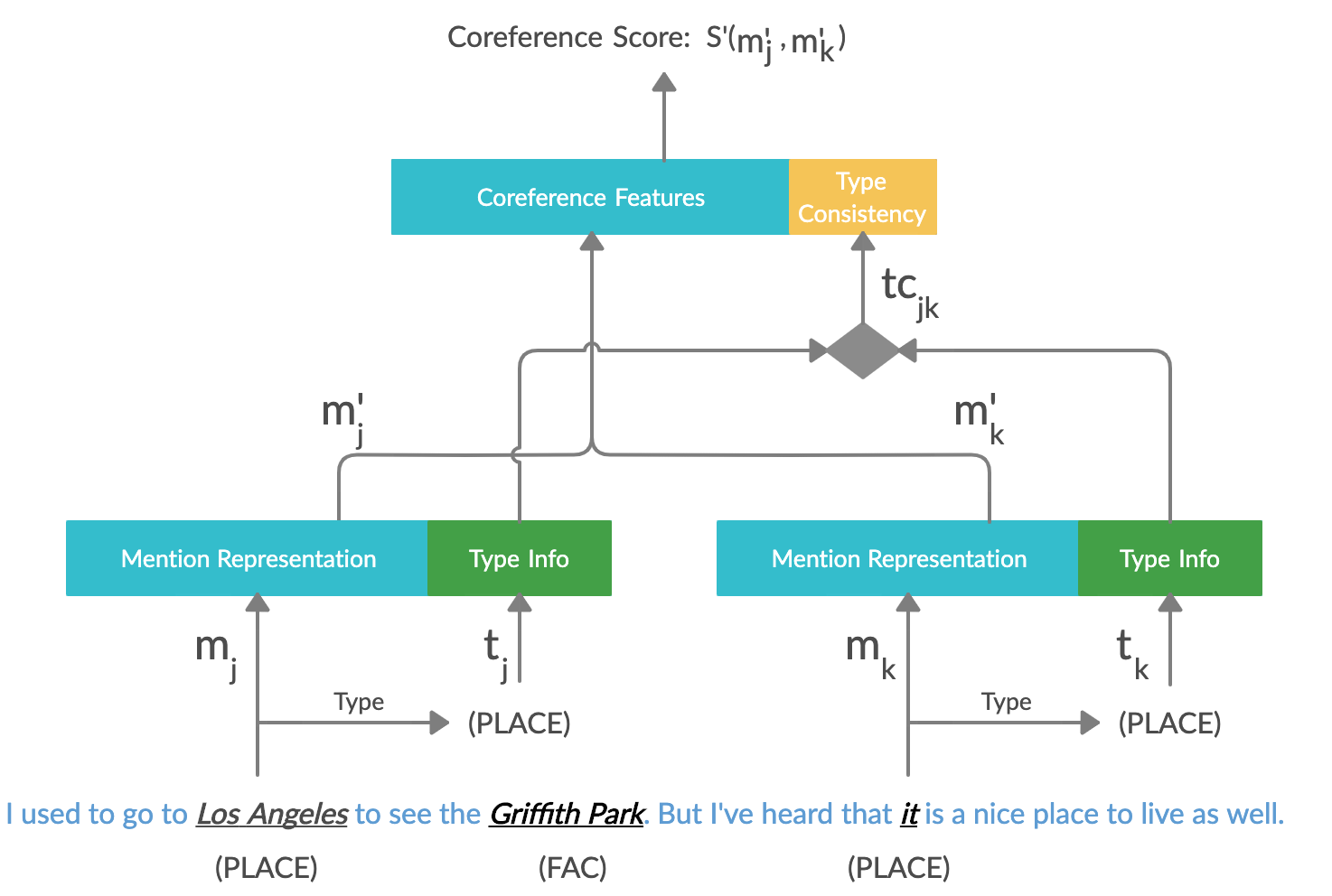}
    \captionsetup{font=small,labelfont=small}
    \caption{We improve~\citet{bamman2020annotated} for entity coreference resolution by incorporating type information at two levels. (1) Type information is concatenated with the mention span representation created by their model; and (2) A consistency check is incorporated that compares the types of two mentions under consideration to calculate the coreference score. Please refer to Section~\ref{sec:model} for details.}
    \label{fig:main_fig}
\end{figure*}

In this paper, we present a proof of concept to portray the benefits of using type information in neural approaches for CR. 
Named entities are generally divided generically (e.g. person, organization etc.) or in a domain-specific manner (e.g. symptom, drug, test etc.). In this work, we consider CR datasets that contain generic entity-types. 
One challenge is that the different corpora do not utilize the same set of type tags.  For example, OntoNotes includes 18 types while EmailCoref includes only 4.  Thus, we evaluate the performance of the proposed modeling approach on each dataset both with the set of type tags germaine to the dataset as well as a common set of four basic types (person, org, location, facility) inspired from research on Named Entity Recognition (NER)~\cite{tjong2002conll, tjong2003introduction}.

Our motivation is similar to 
\cite{durrett-klein-2014-joint}, which used a structured CRF with hand-curated features to jointly-model the tasks of CR, entity typing, and entity linking. Their joint architecture showed an improved performance on CR over the independent baseline. 
However, our work differs from there's as we show the benefits of entity-type information in neural models that use contextualized representations like BERT~\cite{peters2018bert}. Some prior art~\cite{petroni2019language,roberts2020much} argues that contextualized embeddings implicitly capture facts and relationships between real-world entities. However, in this work, we empirically show that access to explicit knowledge about entity-types benefits neural models that use BERT for CR. We show a consistent improvement in performance on four different coreference datasets from varied domains.

Our contribution is that we evaluate the impact of the introduction of type information in neural entity coreference at two different levels of granularity (which we refer to as original vs common), demonstrating their utility both in the case where gold standard type information is available, and the more typical case where it is predicted.

	\section{Related Work}
\label{sec:rel_work}
\noindent\textbf{Neural Coreference Resolution:}
Recently, neural approaches to coreference~\cite{joshi2020spanbert, joshi2019bert, lee2018higher,lee2017end} have begun to show their prowess. The SOTA models show impressive performance on state-of-the-art datasets like OntoNotes~\cite{onto2012} and GAP~\cite{gap}. The notable architecture proposed by \citet{lee2017end} scores pairs of entity mentions independently and later uses a clustering algorithm to find coreference clusters. On the other hand, \citet{lee2018higher} improve upon this foundation by introducing an approximated higher-order inference that iteratively updates the existing span representation using its antecedent distribution. Moreover, they propose a coarse-to-fine grained approach to pairwise scoring for tackling the computational challenges caused due to the iterative higher-order inference. More recently, \citet{joshi2019bert,joshi2020spanbert} showed that use of contextual representations instead of word-embeddings like GloVe~\cite{pennington2014glove} can further boost the results over and above those just mentioned. Our work offers additional improvement by building on the model proposed in~\citet{bamman2020annotated}, which is based on \citet{lee2017end}, and adds additional nuanced information grounded in semantic types.

\noindent\textbf{Type Information:}
Named Entity Recognition datasets~\cite{tjong2002conll, tjong2003introduction, bioner} often group entity mentions into different types (or categories) depending on the domain and the potential downstream applications of the corpus. For example, the medical corpus used in the i2b2 Challenge 2010~\cite{i2b2} annotates domain-specific types like \textit{problem}, \textit{test}, \textit{symptom} etc., whereas, a more general-domain dataset like CoNLL-2002~\cite{tjong2002conll} uses generic types like \textit{person}, \textit{organization}, and \textit{location}. Type information as a predictive signal has been shown to be beneficial for NLP tasks like relation extraction~\cite{soares2019matching} and entity-linking~\cite{type_entitylinking}. It affords some level of disambiguation, which assists models with filtering out some incorrect predictions in order to increase the probability of a correct prediction. In this work, we evaluate the benefits of using explicit type information for CR. We show that a model that leverages entity types associated with the anaphoric/ antecedent mentions significantly reduces the problem of type inconsistency in the output coreference clusters and thus improves the overall performance of the neural baseline on four datasets.

\noindent\textbf{Type Information for CR:}
Multiple prior works have shown type-information to be a useful feature for shallow coreference resolution classifiers~\cite{soon2001machine, bengtson2008understanding, ponzetto2006exploiting,haghighi2010coreference,durrett-klein-2014-joint}.~\cite{soon2001machine} take the most frequent sense for each noun in WordNet as the semantic class for that noun and use a decision-tree for pairwise classification of whether two samples co-refer each other. \cite{bengtson2008understanding} use a hypernym tree to extract the type information for different common nouns, and compare the proper names against a predefined list to determine if the mention is a person. They, then, pass this and many other features (like distance, agreement, etc.) through a regularized average perceptron for pairwise classification. 
This paper expands on these studies to show that entity-type information is also beneficial for neural models that use contextualized representations like BERT~\cite{peters2018bert}, which have been argued to implicitly capture facts and relationships between real-world entities~\cite{petroni2019language,roberts2020much}. 
	\section{Model}
\label{sec:model}
In this section, we explain how we introduce type information into a neural CR system.
\subsection{Baseline}
\label{sec:baseline}
We use the model proposed in~\citet{bamman2020annotated} as our baseline. The model gives state-of-the-art scores on the LitBank corpus~\cite{bamman2020annotated} and is an end-to-end mention ranking system based on~\citet{lee2017end}, which has shown competitive performance on the OntoNotes dataset. However, this model differs from~\citet{lee2017end} as it uses BERT embeddings, omits author and genre information, and only focuses on the task of mention-linking. Since our main goal is to evaluate the benefits of type information, we too separate mention-linking from mention-identification and only show results computed over gold-standard mentions. This controls for the effects of the mention-identification module's performance on our experiments. 
Impact of type-information incorporation in the real-world end-to-end CR setting (mention identification + linking) is left as future work.

The BERT embeddings for each token $i$ are passed through a bi-directional LSTM ($x_i$). To represent a mention $m$ with start and end positions $s, e$ respectively, $x_{s}$, $x_{e}$, attention over ${x_{s}, ..., x_{e}}$, and features to represent the width ($wi$) and inclusion within quotations ($qu$) are concatenated. 
\begin{equation}
    m = [x_{s}; x_{e}; Att(x_{s}, .., x_{e}); wi; qu]
    \label{eq:1}
\end{equation}
Finally, given the representation of two mentions $m_j$ and $m_k$, their coreference score $\mathbf{S}(m_j, m_k)$ is computed by concatenating $m_j$, $m_k$, $m_j \odot m_k$, distance ($d$) between the mentions and whether one mention is nested ($n$) within the other, which are then passed through fully-connected layers ($\mathbf{FC}$).
\begin{equation}
    \mathbf{S}(m_j, m_k) = \mathbf{FC}([m_j; m_k; m_j \odot m_k; d; n])
    \label{eq:2}
\end{equation}
We refer the reader to \cite{bamman2020annotated, lee2017end} for more details about the architecture.

\subsection{Entity Type Information}
We improve the above model by including entity-type information on two levels (Figure~\ref{fig:main_fig}). First, we concatenate the entity-type $t$ of the mention to $m$ (in Eq.~\ref{eq:1}) to improve the mention representation. 
\begin{equation}
    m' = [m; t]
    \label{eq:3}
\end{equation}
This allows the model access to the entity type of the mention as an additional feature. We call this \textbf{+ET-self}. Second, to check the type consistency (softly) between any two mentions under consideration as possibly coreferent, we append a feature ($tc$) in Eq.~\ref{eq:2}, which takes the value 0 if both mentions have the same type, and 1 otherwise. For example, in Figure~\ref{fig:main_fig}, since \textit{Los Angeles} and \textit{it} have the same entity-type PLACE, $tc_{jk} = 0$.
\begin{equation}
    \mathbf{S'}(m'_j, m'_k) = \mathbf{FC}([m'_j; m'_k; m'_j \odot m'_k; d; n; tc_{jk}])
    \label{eq:4}
\end{equation}
This part of the approach is referred to as \textbf{+ET-cross} throughout the remainder of the paper. We decide against the use of a hard consistency check (which would filter out mentions which do not have the same type) as it might not generalize well to bridging anaphora~\cite{clark1975bridge} where the anaphor refers to an object that is associated with, but not identical to, the antecedent~\cite{poesio2018anaphora}. In such cases, the type of the anaphora and its antecedent may not match. Finally, our architecture combines both components together as \textbf{+ET} (ET = ET-self + ET-cross). 
	\section{Datasets}
\label{sec:data}
We gauge the benefits of using entity-type information on the four datasets discussed below.

\noindent\textbf{LitBank.} This dataset~\cite{bamman2020annotated} contains coreference annotations for 100 literary texts.\footnote{https://github.com/dbamman/lrec2020-coref} This dataset limits the markable mentions to six entity-types, where majority of the mentions ($83.1\%$) point to a person. 

\noindent\textbf{EmailCoref.} This dataset~\cite{dakle2020study} comprises of 46 email threads with a total of 245 email messages.\footnote{https://github.com/paragdakle/emailcoref} Similar to LitBank, it considers a mention to be a span of text that refers to a real-world entity. In this work, we filter out pronouns that point towards multiple entities in the email (e.g. we, they) thus only focusing on singular mentions.

\noindent\textbf{Ontonotes.}
From this multi-lingual dataset, we evaluate on the subset (english) from OntoNotes that was used in the CoNLL-2012 shared task~\cite{onto2012}.\footnote{https://catalog.ldc.upenn.edu/LDC2013T19} It contains 2802 training, 343 development, and 348 test documents. The dataset differs from LitBank in its annotation scheme with the biggest difference being the fact that it does not annotate singletons. 

It contains annotations for 18 different entity-types. However, unlike LitBank and EmailCoref, not all mentions have an associated entity-type. For example, none of the pronoun mentions are given a type even if they act as anaphors to typed entities. We partially ameliorate this issue by extracting gold coreference clusters that contain at least one typed mention and assigning the majority type in that cluster to all of its elements.
For example, in Figure~\ref{fig:main_fig}, if \textit{Los Angeles} is typed PLACE, and \textit{it} is in the gold coreference cluster of \textit{Los Angeles} (no other element in the cluster), then \textit{it} is also assigned the type PLACE.

\noindent\textbf{WikiCoref.} This corpus, released by~\cite{ghaddar2016wikicoref}, comprises 30 documents from wikipedia annotated for coreference resolution.\footnote{http://rali.iro.umontreal.ca/rali/?q=en/wikicoref} 
The annotations contain additional metadata, like the associated freebase rdf link for each mention (if available). We use this rdf entry to extract the mention's entity types from freebase dump. Mentions that do not get any type are marked \textit{NA}.
The first 24 documents are chosen for training, the next 3 for development, and the rest for testing.
\begin{table}[t]
    \centering
    \resizebox{\linewidth}{!}{
        \begin{tabular}{ccL{6cm}}
        \toprule
            \textbf{Dataset} & \textbf{\#Types} & \textbf{Categories}\\
        \midrule
            \textbf{LitBank} & 6 & PER, LOC, FAC, GPE, VEH, ORG\\
        \midrule
            \textbf{EmailCoref} & 4 & PER, ORG, LOC, DIG\\
        \midrule
            \textbf{WikiCoref} & 8 & ORG, PER, CORP, EVENT, PLACE, THING, OTHER, NA\\
        \midrule
            \textbf{OntoNotes} & 19 & ORG, WOA, LOC, CARD, EVENT, NORP, GPE, DATE, PER, FAC, QUANT, ORD, TIME, PROD, PERC, MON, LAW, LANG, NA\\
        \midrule
        \midrule
            \textbf{Common} & 5 & PER, ORG, LOC, FAC, OTHER \\
        \bottomrule
        \end{tabular}
    }
    \captionsetup{font=small,labelfont=small}
    \caption{Type statistics for corpora used in this study.}
    \label{tab:types}
\end{table}

\noindent The above-discussed datasets differ in the number as well as the categories of entity-types they originally annotate (Table~\ref{tab:types}). Apart from a common list of types (like PER, ORG, LOC), they also include corpus-specific categories like DIGital (EmailCoref), MONey, and LANG (OntoNotes). We carry out experiments with two sets of types -- original and common -- for each dataset.  The common set of types include the following 5 categories: PER, ORG, LOC, FAC, OTHER.

	\section{Experiments and Results}
\label{sec:results}
In this section, we provide the results of our empirical experiments.

\noindent\textbf{Evaluation Metrics:}
We convert all three datasets into the CoNLL 2012 format and report the F1 score for MUC, $\text{B}^3$, and CEAF metrics using the CoNLL-2012 official scripts. The performances are compared on the average F1 of the above-mentioned metrics.

For EmailCoref, OntoNotes, and WikiCoref, we report the mean score of 5 independent runs of the model with different seeds. Whereas, for LitBank, we present the 10-fold cross-validation results.\footnote{Hyperparameter values are provided in Appendix~\ref{app:hyper}.}

\subsection{Performance with Original Types}
\label{sec:all_types}
In order to establish an upper bound for improvement through introduction of type information, our first experiment leverages the original list of entity-types annotated in different corpora (\textbf{+ ET (orig)}), using the gold standard labels for types. Inclusion of entity-type information improves over the baseline for all CR datasets. Table~\ref{tab:main_result} presents the performance of the baseline model and the model with entity-type information. 
\begin{table}[t]
	\centering
	\resizebox{\linewidth}{!}{
	\begin{tabular}{lccccc}
		\toprule
		\textbf{Model} & \textbf{$\text{B}^3$} & \textbf{MUC} & \textbf{CEAFE} & \textbf{Avg. F1} & \textbf{\#IC}\\
		\midrule
		\multicolumn{6}{c}{\textbf{LitBank~\cite{bamman2020annotated}}}\\
		\midrule
		\textbf{Baseline} & 72.7 & 88.5 & 76.7 & 79.30 & 26\\
		\textbf{+ ET (orig)} & \textbf{74.1} & \textbf{89.2} & \textbf{77.5} & \textbf{80.26} & \textbf{5}\\
		\textbf{+ ET (com)} & 73.3 & 89.1 & 77.5 & 79.97 & 13\\
		\midrule
		\multicolumn{6}{c}{\textbf{EmailCoref~\cite{dakle2020study}}}\\
		\midrule
		\textbf{Baseline} & 72.8 & 84.5 & 62.7 & 73.33 & 9\\
		\textbf{+ ET (orig)} & \textbf{74.9} & 86.7 & \textbf{66.9} & \textbf{76.17} &\textbf{0}\\
		\textbf{+ ET (com)} & 74.8 & \textbf{86.8} & 66.7 & \textbf{76.10} & 2\\
		\midrule
		\multicolumn{6}{c}{\textbf{WikiCoref~\cite{ghaddar2016wikicoref}}}\\
		\midrule
		\textbf{Baseline} & 70.7 & 80.0 & 54.7 & 68.45 & 133\\
		\textbf{+ ET (orig)} & 73.0 & \textbf{82.8} & \textbf{58.3} & \textbf{71.35} & \textbf{70}\\
		\textbf{+ ET (com)} & \textbf{73.4} & 80.7 & 55.1 & 69.71 & 94\\
		\midrule
		\multicolumn{6}{c}{\textbf{OntoNotes~\cite{onto2012}}}\\
		\midrule
		\textbf{Baseline} &82.3 & 90.8 & 77.1 & 83.36 & 60\\
		\textbf{+ ET (orig)} & \textbf{84.4} & \textbf{92.9} & \textbf{79.9} & \textbf{85.76} & \textbf{44}\\
		\textbf{+ ET (com)} & 84.6 & 92.6 & 79.8 & 85.68 & 46\\
		\bottomrule
	\end{tabular}
	}
	\captionsetup{font=small,labelfont=small}
	\caption{CR results with entity-type information (gold mentions). \textbf{IC} = \textbf{I}mpure \textbf{C}lusters.}
	\label{tab:main_result}
\end{table}

We find that entity-type information gives a boost of 0.96 Avg. F1 ($p < 0.01$) on LitBank which is the new state-of-the-art score with gold-mentions. This suggests that type information is helpful for CR on LitBank despite the heavily skewed distribution of entity-types in this corpus. 
Similarly, type information also benefits EmailCoref and WikiCoref resulting in an absolute improvement of 1.67 and 2.9 Avg. F1 points respectively ($p < 0.01$).
We also see a 2.4 Avg. F1 improvement ($p < 0.01$) on OntoNotes, the largest dataset in this study. This suggests that explicit access to type information is beneficial all over the board, despite the use of contextual representations which have been claimed to model real-world facts and relationships~\cite{petroni2019language}.

\noindent\textbf{Ablation Results:}
\label{sec:ablation_results}
To understand the contribution of the inclusion of type information to improve mention representation (+ET-self) and type consistency check between candidate mentions (+ET-cross), we perform an ablation study (Table~\ref{tab:ablation}). We find that both components consistently provide significant performance boosts over the baseline. However, their combination (+ET) performs the best across all datasets.

\subsection{Performance with Common Types}
\label{sec:common_types}
The previous experiment leverages the original entity-types assigned by dataset annotators. Due to the differences in domain and annotation guidelines among these datasets, the annotators introduce several domain-specific entity types (e.g. DIGital, Work Of Art etc.) apart from the common four (PERson, ORGanization, LOCation, FACility) that are often used in the Named Entity Recognition literature~\cite{tjong2002conll}. The former can prove to be much more difficult to obtain/ learn due to dearth of relevant data. Therefore, to assess the worth of using a common entity-type list for all datasets, we map the original types (Table~\ref{tab:types}) to the above-mentioned four common types.
\footnote{We provide the mapping between the original types and the common types in Appendix~\ref{app:type_map}.} 
Categories that do not map to any common type are assigned \textit{Other}. \textbf{+ET (com)} rows in Table~\ref{tab:main_result} show the results for this experiment. Models trained with common types as features perform worse than \textbf{+ET (orig)} which was expected as several original types are now clubbed into a single category (e.g. LAW -> OTHER, LANG -> OTHER) thus somewhat reducing the effectiveness of the feature. One surprising observation is the small difference between the performance on OntoNotes dataset, despite the fact that the number of type categories reduce from 18 + Other (\textbf{+ET (orig)}) to 4 + Other (\textbf{+ET (com)}). This could either be because (1) the entities with corpus-specific types occur less frequently in Ontonotes, or (2) the baseline model does a good job in resolving them. Further research is required to understand this case which is out of scope for this work.
\begin{table}[t]
	\centering
	\resizebox{\linewidth}{!}{
	\begin{tabular}{ccccc}
		\toprule
		\textbf{Dataset} & \textbf{Baseline} & \textbf{+ ET-self} & \textbf{+ ET-cross} & \textbf{+ ET (orig)}\\
		\midrule
		\textbf{LitBank} & 79.30 & 79.97 & 79.76 & \textbf{80.26} \\
		\textbf{EmailCoref} & 73.33 & 73.98 & 74.32 & \textbf{76.17} \\
		\textbf{WikiCoref} & 68.45 & 70.89 & 70.71 & \textbf{71.35} \\
		\textbf{OntoNotes} & 83.36 & 85.55 & 85.69 & \textbf{85.76} \\
		\bottomrule
	\end{tabular}
	}
	\captionsetup{font=small,labelfont=small}
	\caption{Ablation results (\textbf{ET} = \textbf{ET-self} + \textbf{ET-cross}).}
	\label{tab:ablation}
\end{table}

\subsection{\# Impure Clusters (\#IC)}
\label{sec:clust_impurity}
Our hypothesis around the use of entity-types was to provide additional information to the model that could be leveraged to minimize errors due to type mismatch in CR. To evaluate if the F1 score improvements achieved by \textbf{+ET} models are because of fewer type mismatch errors, we report the number of coreference clusters detected by the model that contain at least one element with a type that is different from the others in the cluster. Since all of the datasets used in this work only consider identity coreferences~\cite{recasens2011identity} - with potentially varied definitions of identity~\cite{bamman2020annotated, onto2012} - where the mention is a linguistic "re-packaging" of its antecedent, this measure makes sense. As shown in Table~\ref{tab:main_result}, 
the
models that score lower on the impurity measure get a higher Avg F1. This suggests that the aggregate performance improvements are at least partly due to the better mention-mention comparison in \textbf{+ET} systems.
	\begin{figure}
    \centering
    \includegraphics[width = 0.48\textwidth]{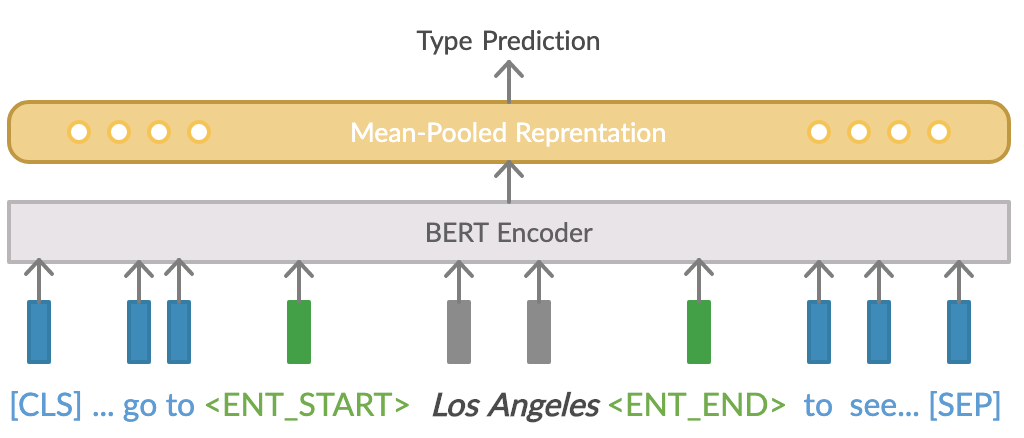}
    \caption{Type-prediction model.}
    \label{fig:type_pred_fig}
\end{figure}

\begin{table*}[t]
	\centering
	\resizebox{\textwidth}{!}{
	\begin{tabular}{c|c|c|cccccc}
		\toprule
		\textbf{Dataset} & \textbf{Macro F1} & \multicolumn{6}{c}{\textbf{Accuracy}} \\
    		 \textbf{(Types)}& \textbf{All} & \textbf{All} & \textbf{PRP (dem.)} & \textbf{PRP (pers.)} & \textbf{NP (len = 1)} & \textbf{NP (len = 2)} & \textbf{NP (len > 2)}\\
    	\midrule
    		\textbf{LitBank (orig)} & 84.0  & 97.0  & 70.5 (207) & 99.75 (6127) & 96.5 (6477) & 95.8 (5439) & 94.03 (4443) \\ 
    		\textbf{LitBank (com)} & 87.0 & 97.0  & 72.5 (207) & 99.8 (6127) & 97.2 (6477) & 96.7 (5439) & 94.8 (4443) \\
		\midrule
    		\textbf{EmailCoref (orig)} & 83.0 & 91.0 & 77.1 (35) & 83.8 (482) & 93.6 (1635) & 91.3 (484) & 94.1 (1058) \\
    		\textbf{EmailCoref (com)} & 85.0 & 92.0 & 65.7 (35) & 84.6 (482) & 93.5 (1635) & 90.3 (484) & 95.5 (1058) \\
    	\midrule
    		\textbf{WikiCoref (orig)} & 38.0 & 61.0 & 53.3 (75) & 59.4 (340) & 58.9 (1531) & 55.1 (1223) & 54.5 (1218)  \\
    		\textbf{WikiCoref (com)} & 45.0 & 74.0 & 82.7 (75) & 56.5 (340) & 64.3 (1531) & 69.2 (1223) & 68.7 (1218) \\
    	\midrule
		\textbf{OntoNotes (orig)} & 74.0 & 95.0 & 80.2 (1698) & 94.6 (8682) & 96.2 (21589) & 95.5 (13297) & 94.0 (11951) \\
		\textbf{OntoNotes (com)} & 90.0 & 96.0 & 84.5 (1698) & 94.3 (8682) & 96.9 (21589) & 96.4 (13297) & 95.0 (11951)\\
		\bottomrule
	\end{tabular}
	}
	\captionsetup{font=small,labelfont=small}
	\caption{Performance of our BERT-based model for type-prediction. 
	The last five columns show the accuracy (\# samples) on demonstrative and personal pronouns, and noun phrases of different lengths. 
	\textbf{PRP (dem.)} = Demonstrative Pronouns (this, that, it, these, those), \textbf{PRP (pers.)} = Personal Pronouns (she, he, they, me, you, we). 
	The scores for WikiCoref and OntoNotes do not include mentions without an associated gold type (\textit{NA}).}
	\label{tab:pred_perf}
\end{table*}
\section{Predicted Types}
Results shown in the previous section assume the presence of gold standard types during training as well as inference, which is often impractical in the real-world. Most of the new samples that a CR model would encounter would not include type information about the candidate mentions. Therefore, we set up an additional experiment to gauge the benefits of type information using predicted types. We introduce a baseline approach to infer the type of the mentions and then use these predictions in the +ET models, in place of the gold types, for coreference resolution.

\subsection{Type Prediction Model}
\label{sec:type_pred_model}
Given the mention and its immediate context, i.e. the sentence it occurs in ($S = ..., c_{-2}, c_{-1}, \mathbf{e_{1}, e_{2}, ..., e_{n}}, c_{1}, c_{2}, ...$), we add markers \url{<ENT\_START>}/ \url{<ENT\_END>} before/ after the beginning/ ending of the mention in the sentence. The new sequence ($S'= ..., c_{-2}, c_{-1},$ \url{<ENT\_START>}, $\mathbf{e_{1}, e_{2}, ..., e_{n},}$ \url{<ENT\_END>}, $c_{1}, c_{2}, ...$) is tokenized using BERT tokenizer and passed through the BERT encoder. The output from which is then mean-pooled and passed through a fully-connected layer for classification. This architecture is motivated from~\cite{soares2019matching} who show that adding markers around entities before passing the sentence through BERT performs better for relation extraction.

\subsection{Experiments and Results}
\label{sec:pred_exp}
\noindent\textbf{Type Prediction: }
Our final evaluation of the use of types in coreference is perhaps the most important one as it uses predicted types rather than annotated types, thus demonstrating that the benefits can be achieved in practice.  Here we use the Type Prediction Model described just above.  We limit the length of the input sequence to 128 tokens and use BERT-base-cased model for our type-prediction experiments.
We perform a five-fold cross-validation to predict the type for each mention in the dataset. Since all four datasets suffer from class-imbalance, we report both Macro F1 score as well as the accuracy for the model. The model is trained for 20 epochs, with early-stopping (patience = 10), and is fine-tuned on the development set for Macro F1 to give more importance to minority type categories. We do not consider NA as a separate class during type prediction for WikiCoref and OntoNotes. For evaluation of our type-prediction model, we ignore the mentions that do not have an associated gold type (NA) from the final numbers in Table~\ref{tab:pred_perf}.

As shown, our model performs well on LitBank, EmailCoref, and Ontonotes  due to their favorable size in terms of training samples for the BERT-based type predictor. WikiCoref, however, proves more challenging as the model only manages $38.0$ Macro F1 points with original (orig) types and $45.0$ with common types (com), portraying its lack of ability to learn minority type categories with less data. Furthermore, our model finds it easier to predict the common (com) set of types for each dataset as combining multiple corpus-specific types into one partially alleviates the problem of class-imbalance. In line with our expectation, the largest improvement due to common types is seen for OntoNotes where the problem reduces from an 18-way classification to a 5-way classification.

\begin{table}[t]
	\centering
	\resizebox{\linewidth}{!}{
	\begin{tabular}{cccc}
		\toprule
		\textbf{Dataset} & \textbf{Baseline} & \textbf{+ ET-pred (orig)} & \textbf{+ ET-pred (com)}\\
		\midrule
		\textbf{LitBank} & 79.30 & \textbf{79.58} & 79.40\\ 
		\textbf{EmailCoref} & 73.33 & \textbf{74.20} & \textbf{74.18}\\ 
		\textbf{WikiCoref} & 68.45 & 68.37 & 68.62\\
		\textbf{OntoNotes} & 83.36 & \textbf{84.02} & 83.65 \\
		\bottomrule
	\end{tabular}
	}
	\captionsetup{font=small,labelfont=small}
	\caption{CR results with predicted types. Numbers in \textbf{bold} are significantly better ($p < 0.01$) than the baseline.}
	\label{tab:pred_ablation}
\end{table}

\begin{table*}[!htb]
    \centering
    \resizebox{0.75\textwidth}{!}{
    \begin{tabular}{cccl}
    \toprule
        \textbf{Genre} & \textbf{Baseline} & \textbf{+ET (orig)} & \textbf{Most Frequent Entity-Types (Ratio)} \\
    \midrule
        \textbf{tc} & 81.15 & \textbf{85.97} & PER (0.839), GPE (0.083)\\
    \midrule
        \textbf{bc} & 79.66 & \textbf{83.25} & PER (0.456), GPE (0.277), NORP (0.098), ORG (0.089) \\
    \midrule
        \textbf{nw} & 84.91 & 85.26 & ORG (0.411), GPE (0.232), PER (0.161), DATE (0.132)\\
    \midrule
        \textbf{pt} & 84.7 & 85.66 & -\\
    \midrule
        \textbf{bn} & 84.22 & \textbf{87.46} & PER (0.484), GPE (0.237), ORG (0.132), DATE (0.052)\\
    \midrule
        \textbf{wb} & 80.57 & \textbf{83.52} & PER (0.744), GPE (0.130), ORG (0.092)\\
    \midrule
        \textbf{mz} & 89.21 & \textbf{91.92} & GPE (0.513), PER (0.273), ORG (0.198)\\
    \bottomrule
    \end{tabular}
    }
    \captionsetup{font=small,labelfont=small}
    \caption{Type Distribution (excluding OTHER) for mentions within different Ontonotes genres after clustering based type-propagation. Numbers in \textbf{bold} are significantly better than the baseline ($p < 0.01$). - represents that none of the mentions in the genre are annotated with types.}
    \label{app:ent_dist}
\end{table*}

\noindent\textbf{Coreference Resolution:}
Each mention in the corpus occurs in the test-sets of the five-fold cross-validation type-prediction experiments exactly once. This allows us to infer the type of each mention using the model that is trained on a different subset of the dataset. These inferred types are used in the training and testing of the CR systems in a manner similar to the annotated types. 

Empirically, we found that the above configuration performs better than using the \textbf{+ET} models trained with annotated types and testing with predicted types, as the former exposes the CR models to the noisy types during training thus allowing them to learn weights that take this noise into account. We report the results for both original (\textbf{+ ET-pred (orig)}) and common (\textbf{+ ET-pred (com)}) type categories on each dataset.

Table~\ref{tab:pred_ablation} shows the results for performance of the baseline and the type-informed models on the four datasets, where the types are inferred from the model described in Section~\ref{sec:type_pred_model}. We find that the improvements from type-information persist across LitBank, EmailCoref, and OntoNotes despite the use of predicted types, but, quite expectedly, remain smaller than the improvements from the gold annotated types. Scores on WikiCoref show no significant improvement over the baseline, which could be explained by the poor performance of the type prediction model on this dataset which reduces the potency of the feature for CR.


	\section{Discussion}
\label{sec:discuss}
\subsection{Genres of OntoNotes}
\begin{table*}[ht]
    \centering
    \resizebox{\linewidth}{!}{
    \begin{tabular}{c|L{16.1cm}}
    \toprule
         & \textbf{Example}\\
    \midrule
        Gold & Hi Chris \textbf{(0)} ,
         I \textbf{(1)} will have \{Tamara Utsch\} \textbf{(2)} reply to you \textbf{(0)} on your (0) refund -
         I \textbf{(1)} believe she \textbf{(2)} updated you \textbf{(0)} last week , but we \textbf{(3)} 'll see if there 's new news this week .
         Can you \textbf{(0)} please send me \textbf{(1)} \{your \textbf{(0)} current home address\} \textbf{(4)} so we \textbf{(3)} can send you \textbf{(0)} \{an organizer\} \textbf{(5)} ?
         Thanks ,
         \{Judy Perdomo\} \textbf{(1)}
         \{PricewaterhouseCoopers Calgary\} \textbf{(3)} \\
    \midrule
        Baseline & Hi Chris \textbf{(0)} ,
         I \textbf{(1)} will have \{Tamara Utsch\} \textbf{(2)} reply to you \textbf{(0)} on your (0) refund -
         I \textbf{(1)} believe she \textbf{(2)} updated you \textbf{(0)} last week , but we \textbf{(3)} 'll see if there 's new news this week .
         Can you \textbf{(0)} please send me \textbf{(1)} \{your \textbf{(0)} current home address\} \textbf{(4)} so we \textbf{(3)} can send you \textbf{(0)} \{an organizer\} {\color{red} \underline{\emph{(4)}}} ?
         Thanks ,
         \{Judy Perdomo\} \textbf{(1)}
         \{PricewaterhouseCoopers Calgary\} {\color{red} \underline{\emph{(6)}}} \\
    \midrule
        + ET (orig) & Hi Chris \textbf{(0)} ,
         I \textbf{(1)} will have \{Tamara Utsch\} \textbf{(2)} reply to you \textbf{(0)} on your (0) refund -
         I \textbf{(1)} believe she \textbf{(2)} updated you \textbf{(0)} last week , but we \textbf{(3)} 'll see if there 's new news this week .
         Can you \textbf{(0)} please send me \textbf{(1)} \{your \textbf{(0)} current home address\} \textbf{(4)} so we \textbf{(3)} can send you \textbf{(0)} \{an organizer\} \textbf{(5)} ?
         Thanks ,
         \{Judy Perdomo\} \textbf{(1)}
         \{PricewaterhouseCoopers Calgary\} \textbf{(3)} \\
    \midrule
        + ET (com) & Hi Chris \textbf{(0)} ,
         I \textbf{(1)} will have \{Tamara Utsch\} \textbf{(2)} reply to you \textbf{(0)} on your (0) refund -
         I \textbf{(1)} believe she \textbf{(2)} updated you \textbf{(0)} last week , but we \textbf{(3)} 'll see if there 's new news this week .
         Can you \textbf{(0)} please send me \textbf{(1)} \{your \textbf{(0)} current home address\} \textbf{(4)} so we \textbf{(3)} can send you \textbf{(0)} \{an organizer\} \textbf{(5)} ?
         Thanks ,
         \{Judy Perdomo\} \textbf{(1)}
         \{PricewaterhouseCoopers Calgary\} \textbf{(3)} \\
    \midrule
        + ET-pred (orig) & Hi Chris \textbf{(0)} ,
         I \textbf{(1)} will have \{Tamara Utsch\} \textbf{(2)} reply to you \textbf{(0)} on your (0) refund -
         I \textbf{(1)} believe she \textbf{(2)} updated you \textbf{(0)} last week , but we \textbf{(3)} 'll see if there 's new news this week .
         Can you \textbf{(0)} please send me \textbf{(1)} \{your \textbf{(0)} current home address\} \textbf{(4)} so we \textbf{(3)} can send you \textbf{(0)} \{an organizer\} \textbf{(5)} ?
         Thanks ,
         \{Judy Perdomo\} \textbf{(1)}
         \{PricewaterhouseCoopers Calgary\} {\color{red} \underline{\emph{(6)}}} \\
    \bottomrule
    \end{tabular}
    }
    \captionsetup{font=small,labelfont=small}
    \caption{An example email from EmailCoref corpus. Numbers in round brackets denote the cluster number of the mention. {\color{red} Red} denotes incorrect predictions.}
    \label{tab:example}
\end{table*}
Table~\ref{app:ent_dist} shows the most frequently occurring entity-types for each of the genres in OntoNotes. In line with our intuition, we find that enity-type information helps the baseline in \textit{bc}, \textit{bn}, \textit{wb}, and \textit{mz} genres which have less skew in their entity-type distribution. Genres like \textit{bc}, \textit{bn} and \textit{wb}, although dominated by PER entities, contain a substantial minority of other entity-types like ORG and GPE. Along the same lines, \textit{mz} contains a majority of GPE entities but also enough entities with type PER and ORG to make type information a potentially useful feature for CR. However, two exceptions to this are the improved performance of \textbf{+ET (orig)} on \textit{tc} (highest skew) and no significant improvement on \textit{nw} (lowest skew). These findings prompt further research in the future.
\subsection{Type Prediction: PRP vs NP}
Entity coreference in discourse often takes the surface form of pronouns (PRP) (like she, they, that, it etc.) or noun phrases (NP) (like LA, John's brother etc.) In Table~\ref{tab:pred_perf}, we compare the performance of our type prediction model on different types of pronouns, and noun phrases of varying length. We find that the model does well in predicting types for personal pronouns (PRP (pers.)) like \textit{she}, \textit{he} and noun phrases (NP). However, it consistently underperforms on demonstrative pronouns (PRP (dem.)) like \textit{this}, \textit{that}, and \textit{it} across all datasets. 

This reduced performance could be due to the fact that demonstrative pronouns do not contain any signal about the type of the entity they refer to. Therefore, the type prediction model has to solely rely on the context to make that decision. However, this is not the case with PRPs (pers.) and NPs where the mention string is usually a strong indicator of the type. This problem is worsened by the imbalance due to the small presence of PRP (dem.) mentions in difference CR datasets. Since, the model does not encounter enough PRPs (dem.), it might not be able to learn to give high importance to context in these cases.

This could be partially alleviated by creating a separate type-prediction path for PRP (dem.) where the mention span is masked before it is passed through the model. A model that is trained with masked mentions would focus more on the context for type prediction and thus could lead to better performance on PRPs (dem.).

One could also experiment with training the type-prediction model on all of the mentions across the four datasets. The common list of types introduced in this work would allow for the creation of a larger training-set that includes mentions from multiple corpora (including external NER datasets) which could provide enough signal for the model to better learn the common types for PRPs (dem.). 

Both these approaches could further boost the results for CR with predicted entity-types, ultimately, reducing the gap between the scores in Table~\ref{tab:main_result} and~\ref{tab:pred_ablation}. However, they are left as future work as they are out of scope for this paper.

\subsection{Case Study}
Table~\ref{tab:example} provides an excerpt of an email from EmailCoref corpus. As shown, the baseline model predicts the coreference clusters for \textit{an organizer} (DIG) and \textit{PricewaterhouseCoopers Calgary} (ORG) incorrectly. For the former, the model mistakes it as a reference to \textit{your current home address} (LOC) which is corrected by the entity-type aware models. For the latter, the baseline considers \textit{PricewaterhouseCoopers Calgary (PCC)} as part of a new coreference cluster, even though it refers to the organization of the email's sender which was previously referred to as \textit{we} in the email. Models with access to gold type information (+ET (orig) and +ET (com)) are able to make that connection. 

+ET-pred (orig), however, is unable to cluster \textit{PCC} correctly which could be due to the fact that the type-prediction model incorrectly classifies the type of \textit{we} as PER rather than ORG. This could lead to the CR model considering \textit{PCC} (ORG) as a new entity in the discourse rather than a postcedent of \textit{we}. This example demonstrates that sentence-level context might not be sufficient in some cases for mention type-disambiguation. We intend to experiment with models that capture long-term context and leverage external knowledge in the future.

	\section{Conclusion}
\label{sec:conc}
In this work, we show the importance of using entity-type information in neural coreference resolution (CR) models with contextualized embeddings like BERT. Models which leverage type information, annotated in the corpus, substantially outperform the baseline on four CR datasets by reducing the number of type mismatches in detected coreference clusters. Since, these datasets vary in number and categories of the types they define, we also experiment with mapping the original corpus types to four common types (PER, ORG, LOC, FAC) based on previous NER research that can be 
learnt more easily through large NER datasets. Models which use these common types perform slightly worse than original types but still show significant improvements over the baseline systems. 

The presence of gold standard types during CR inference is unlikely in practice. Therefore, we propose a model that infers the type of a mention given the mention span and its immediate context to use along side the proposed CR approach. In our evaluation, we find that using types predicted by our model for CR still performs significantly better than the baseline, thus offering stronger evidence that type information holds the potential for practical improvements for CR.




	\section*{Acknowledgements}
	We thank the anonymous reviewers for
    their insightful comments. We are also grateful to the members of the TELEDIA group at LTI, CMU for the invaluable feedback. This work was funded in part by Dow Chemical, and Microsoft.
	
	\bibliographystyle{acl_natbib}
	\bibliography{references}

\begin{thebibliography}{37}
\expandafter\ifx\csname natexlab\endcsname\relax\def\natexlab#1{#1}\fi

\bibitem[{Bamman et~al.(2020)Bamman, Lewke, and Mansoor}]{bamman2020annotated}
David Bamman, Olivia Lewke, and Anya Mansoor. 2020.
\newblock An annotated dataset of coreference in english literature.
\newblock In \emph{Proceedings of The 12th Language Resources and Evaluation
  Conference}, pages 44--54.

\bibitem[{Bengtson and Roth(2008)}]{bengtson2008understanding}
Eric Bengtson and Dan Roth. 2008.
\newblock Understanding the value of features for coreference resolution.
\newblock In \emph{Proceedings of the 2008 Conference on Empirical Methods in
  Natural Language Processing}, pages 294--303.

\bibitem[{Chen et~al.(2020)Chen, Wang, Jiang, and Lin}]{type_entitylinking}
Shuang Chen, Jinpeng Wang, Feng Jiang, and Chin-Yew Lin. 2020.
\newblock Improving entity linking by modeling latent entity type information.
\newblock \emph{arXiv preprint arXiv:2001.01447}.

\bibitem[{Clark(1975)}]{clark1975bridge}
H.~H. Clark. 1975.
\newblock Bridging.
\newblock In \emph{Proceedings of TINLAP}.

\bibitem[{Clark and Manning(2016)}]{coref_cl4}
Kevin Clark and Christopher~D Manning. 2016.
\newblock Deep reinforcement learning for mention-ranking coreference models.
\newblock In \emph{Proceedings of the 2016 Conference on Empirical Methods in
  Natural Language Processing}, pages 2256--2262.

\bibitem[{Dakle et~al.(2020)Dakle, Desai, and Moldovan}]{dakle2020study}
Parag~Pravin Dakle, Takshak Desai, and Dan Moldovan. 2020.
\newblock A study on entity resolution for email conversations.
\newblock In \emph{Proceedings of The 12th Language Resources and Evaluation
  Conference}, pages 65--73.

\bibitem[{Dasigi et~al.(2019)Dasigi, Liu, Marasovi\'{c}, Smith, and
  Gardner}]{Dasigi2019Quoref}
Pradeep Dasigi, Nelson~F. Liu, Ana Marasovi\'{c}, Noah~A. Smith, and Matt
  Gardner. 2019.
\newblock Quoref: A reading comprehension dataset with questions requiring
  coreferential reasoning.
\newblock In \emph{Proc. of EMNLP-IJCNLP}.

\bibitem[{Durrett and Klein(2014)}]{durrett-klein-2014-joint}
Greg Durrett and Dan Klein. 2014.
\newblock \href {https://doi.org/10.1162/tacl_a_00197} {A joint model for
  entity analysis: Coreference, typing, and linking}.
\newblock \emph{Transactions of the Association for Computational Linguistics},
  2:477--490.

\bibitem[{Ghaddar and Langlais(2016)}]{ghaddar2016wikicoref}
Abbas Ghaddar and Philippe Langlais. 2016.
\newblock Wikicoref: An english coreference-annotated corpus of wikipedia
  articles.
\newblock In \emph{Proceedings of the Tenth International Conference on
  Language Resources and Evaluation (LREC'16)}, pages 136--142.

\bibitem[{Haghighi and Klein(2010)}]{haghighi2010coreference}
Aria Haghighi and Dan Klein. 2010.
\newblock Coreference resolution in a modular, entity-centered model.
\newblock In \emph{Human Language Technologies: The 2010 Annual Conference of
  the North American Chapter of the Association for Computational Linguistics},
  pages 385--393.

\bibitem[{Hobbs(1978)}]{coref_cl1}
Jerry~R Hobbs. 1978.
\newblock Resolving pronoun references.
\newblock \emph{Lingua}, 44(4):311--338.

\bibitem[{Joshi et~al.(2020)Joshi, Chen, Liu, Weld, Zettlemoyer, and
  Levy}]{joshi2020spanbert}
Mandar Joshi, Danqi Chen, Yinhan Liu, Daniel~S Weld, Luke Zettlemoyer, and Omer
  Levy. 2020.
\newblock Spanbert: Improving pre-training by representing and predicting
  spans.
\newblock \emph{Transactions of the Association for Computational Linguistics},
  8:64--77.

\bibitem[{Joshi et~al.(2019)Joshi, Levy, Zettlemoyer, and Weld}]{joshi2019bert}
Mandar Joshi, Omer Levy, Luke Zettlemoyer, and Daniel~S Weld. 2019.
\newblock Bert for coreference resolution: Baselines and analysis.
\newblock In \emph{Proceedings of the 2019 Conference on Empirical Methods in
  Natural Language Processing and the 9th International Joint Conference on
  Natural Language Processing (EMNLP-IJCNLP)}, pages 5807--5812.

\bibitem[{Lappin and Leass(1994)}]{coref_cl2}
Shalom Lappin and Herbert~J Leass. 1994.
\newblock An algorithm for pronominal anaphora resolution.
\newblock \emph{Computational linguistics}, 20(4):535--561.

\bibitem[{Lee et~al.(2017)Lee, He, Lewis, and Zettlemoyer}]{lee2017end}
Kenton Lee, Luheng He, Mike Lewis, and Luke Zettlemoyer. 2017.
\newblock End-to-end neural coreference resolution.
\newblock In \emph{Proceedings of the 2017 Conference on Empirical Methods in
  Natural Language Processing}, pages 188--197.

\bibitem[{Lee et~al.(2018)Lee, He, and Zettlemoyer}]{lee2018higher}
Kenton Lee, Luheng He, and Luke Zettlemoyer. 2018.
\newblock Higher-order coreference resolution with coarse-to-fine inference.
\newblock In \emph{Proceedings of the 2018 Conference of the North American
  Chapter of the Association for Computational Linguistics: Human Language
  Technologies, Volume 2 (Short Papers)}, pages 687--692.

\bibitem[{Li et~al.(2016)Li, Sun, Johnson, Sciaky, Wei, Leaman, Davis,
  Mattingly, Wiegers, and Lu}]{bioner}
Jiao Li, Yueping Sun, Robin~J Johnson, Daniela Sciaky, Chih-Hsuan Wei, Robert
  Leaman, Allan~Peter Davis, Carolyn~J Mattingly, Thomas~C Wiegers, and Zhiyong
  Lu. 2016.
\newblock Biocreative v cdr task corpus: a resource for chemical disease
  relation extraction.
\newblock \emph{Database}, 2016.

\bibitem[{Mitkov(1999)}]{coref_cl3}
Ruslan Mitkov. 1999.
\newblock \emph{Anaphora resolution: the state of the art}.
\newblock Citeseer.

\bibitem[{Ng(2017)}]{coref_cl5}
Vincent Ng. 2017.
\newblock Machine learning for entity coreference resolution: A retrospective
  look at two decades of research.
\newblock In \emph{Thirty-First AAAI Conference on Artificial Intelligence}.

\bibitem[{Pennington et~al.(2014)Pennington, Socher, and
  Manning}]{pennington2014glove}
Jeffrey Pennington, Richard Socher, and Christopher~D Manning. 2014.
\newblock Glove: Global vectors for word representation.
\newblock In \emph{Proceedings of the 2014 conference on empirical methods in
  natural language processing (EMNLP)}, pages 1532--1543.

\bibitem[{Peters et~al.(2018)Peters, Neumann, Iyyer, Gardner, Clark, Lee, and
  Zettlemoyer}]{peters2018bert}
Matthew Peters, Mark Neumann, Mohit Iyyer, Matt Gardner, Christopher Clark,
  Kenton Lee, and Luke Zettlemoyer. 2018.
\newblock Deep contextualized word representations.
\newblock In \emph{Proceedings of the 2018 Conference of the North American
  Chapter of the Association for Computational Linguistics: Human Language
  Technologies, Volume 1 (Long Papers)}, pages 2227--2237.

\bibitem[{Petroni et~al.(2019)Petroni, Rockt{\"a}schel, Riedel, Lewis, Bakhtin,
  Wu, and Miller}]{petroni2019language}
Fabio Petroni, Tim Rockt{\"a}schel, Sebastian Riedel, Patrick Lewis, Anton
  Bakhtin, Yuxiang Wu, and Alexander Miller. 2019.
\newblock Language models as knowledge bases?
\newblock In \emph{Proceedings of the 2019 Conference on Empirical Methods in
  Natural Language Processing and the 9th International Joint Conference on
  Natural Language Processing (EMNLP-IJCNLP)}, pages 2463--2473.

\bibitem[{Poesio et~al.(2018)Poesio, Grishina, Kolhatkar, Moosavi, Roesiger,
  Roussel, Simonjetz, Uma, Uryupina, Yu et~al.}]{poesio2018anaphora}
Massimo Poesio, Yulia Grishina, Varada Kolhatkar, Nafise~Sadat Moosavi, Ina
  Roesiger, Adam Roussel, Fabian Simonjetz, Alexandra Uma, Olga Uryupina,
  Juntao Yu, et~al. 2018.
\newblock Anaphora resolution with the arrau corpus.
\newblock In \emph{Proceedings of the First Workshop on Computational Models of
  Reference, Anaphora and Coreference}, pages 11--22.

\bibitem[{Ponzetto and Strube(2006)}]{ponzetto2006exploiting}
Simone~Paolo Ponzetto and Michael Strube. 2006.
\newblock Exploiting semantic role labeling, wordnet and wikipedia for
  coreference resolution.
\newblock In \emph{Proceedings of the Human Language Technology Conference of
  the NAACL, Main Conference}, pages 192--199.

\bibitem[{Pradhan et~al.(2012)Pradhan, Moschitti, Xue, Uryupina, and
  Zhang}]{onto2012}
Sameer Pradhan, Alessandro Moschitti, Nianwen Xue, Olga Uryupina, and Yuchen
  Zhang. 2012.
\newblock Conll-2012 shared task: Modeling multilingual unrestricted
  coreference in ontonotes.
\newblock In \emph{Joint Conference on EMNLP and CoNLL - Shared Task}, CoNLL
  ’12, page 1–40, USA. Association for Computational Linguistics.

\bibitem[{Recasens et~al.(2011)Recasens, Hovy, and
  Mart{\'\i}}]{recasens2011identity}
Marta Recasens, Eduard Hovy, and M~Ant{\`o}nia Mart{\'\i}. 2011.
\newblock Identity, non-identity, and near-identity: Addressing the complexity
  of coreference.
\newblock \emph{Lingua}, 121(6):1138--1152.

\bibitem[{Roberts et~al.(2020)Roberts, Raffel, and Shazeer}]{roberts2020much}
Adam Roberts, Colin Raffel, and Noam Shazeer. 2020.
\newblock How much knowledge can you pack into the parameters of a language
  model?
\newblock \emph{arXiv preprint arXiv:2002.08910}.

\bibitem[{Soares et~al.(2019)Soares, FitzGerald, Ling, and
  Kwiatkowski}]{soares2019matching}
Livio~Baldini Soares, Nicholas FitzGerald, Jeffrey Ling, and Tom Kwiatkowski.
  2019.
\newblock Matching the blanks: Distributional similarity for relation learning.
\newblock In \emph{Proceedings of the 57th Annual Meeting of the Association
  for Computational Linguistics}, pages 2895--2905.

\bibitem[{Soon et~al.(2001)Soon, Ng, and Lim}]{soon2001machine}
Wee~Meng Soon, Hwee~Tou Ng, and Daniel Chung~Yong Lim. 2001.
\newblock A machine learning approach to coreference resolution of noun
  phrases.
\newblock \emph{Computational linguistics}, 27(4):521--544.

\bibitem[{Steinberger et~al.(2007)Steinberger, Poesio, Kabadjov, and
  Je{\v{z}}ek}]{coref_summ1}
Josef Steinberger, Massimo Poesio, Mijail~A Kabadjov, and Karel Je{\v{z}}ek.
  2007.
\newblock Two uses of anaphora resolution in summarization.
\newblock \emph{Information Processing and Management}, 6(43):1663--1680.

\bibitem[{Sukthanker et~al.(2020{\natexlab{a}})Sukthanker, Poria, Cambria, and
  Thirunavukarasu}]{coref_survey}
Rhea Sukthanker, Soujanya Poria, Erik Cambria, and Ramkumar Thirunavukarasu.
  2020{\natexlab{a}}.
\newblock Anaphora and coreference resolution: A review.
\newblock \emph{Information Fusion}, 59:139--162.

\bibitem[{Sukthanker et~al.(2020{\natexlab{b}})Sukthanker, Poria, Cambria, and
  Thirunavukarasu}]{sukthanker2020anaphora}
Rhea Sukthanker, Soujanya Poria, Erik Cambria, and Ramkumar Thirunavukarasu.
  2020{\natexlab{b}}.
\newblock Anaphora and coreference resolution: A review.
\newblock \emph{Information Fusion}, 59:139--162.

\bibitem[{Tjong Kim~Sang(2002)}]{tjong2002conll}
Erik~F. Tjong Kim~Sang. 2002.
\newblock \href {https://www.aclweb.org/anthology/W02-2024} {Introduction to
  the {C}o{NLL}-2002 shared task: Language-independent named entity
  recognition}.
\newblock In \emph{{COLING}-02: The 6th Conference on Natural Language Learning
  2002 ({C}o{NLL}-2002)}.

\bibitem[{Tjong Kim~Sang and De~Meulder(2003)}]{tjong2003introduction}
Erik~F Tjong Kim~Sang and Fien De~Meulder. 2003.
\newblock Introduction to the conll-2003 shared task: language-independent
  named entity recognition.
\newblock In \emph{Proceedings of the seventh conference on Natural language
  learning at HLT-NAACL 2003-Volume 4}, pages 142--147.

\bibitem[{Uzuner et~al.(2011)Uzuner, South, Shen, and DuVall}]{i2b2}
{\"O}zlem Uzuner, Brett~R South, Shuying Shen, and Scott~L DuVall. 2011.
\newblock 2010 i2b2/va challenge on concepts, assertions, and relations in
  clinical text.
\newblock \emph{Journal of the American Medical Informatics Association},
  18(5):552--556.

\bibitem[{Webber(1991)}]{webber1991abstract}
Bonnie~Lynn Webber. 1991.
\newblock Structure and ostension in the interpretation of discourse deixis.
\newblock \emph{Language and Cognitive processes}, 6(2):107--135.

\bibitem[{Webster et~al.(2018)Webster, Recasens, Axelrod, and Baldridge}]{gap}
Kellie Webster, Marta Recasens, Vera Axelrod, and Jason Baldridge. 2018.
\newblock Mind the gap: A balanced corpus of gendered ambiguous pronouns.
\newblock \emph{Transactions of the Association for Computational Linguistics},
  6:605--617.

\end{thebibliography}
	
	\setcounter{figure}{0}
	\renewcommand{\thefigure}{A\arabic{figure}}
	
	\setcounter{table}{0}
	\renewcommand{\thetable}{A\arabic{table}}
	\newpage
	\clearpage
	\appendix
	\section*{Appendix}
	\section{Hyperparameters}
\label{app:hyper}

\begin{table}[!ht]
    \centering
    \begin{tabular}{lr}
    \toprule
        \textbf{Hyperparameter} & \textbf{Value} \\
    \midrule
        BERT & base-cased \\
        BERT weights & freeze \\
        BiLSTM hidden dim & 200 \\
        Type embedding size & 20 \\
        FC-layer 1 size & 150 \\
        FC-layer 2 size & 150 \\
        Dropout & 0.2 \\
    \bottomrule
    \end{tabular}
    \captionsetup{font=small,labelfont=small}
    \caption{Hyperparameter values for our model. We refer the reader to \url{https://github.com/dbamman/lrec2020-coref} for the implementation of the baseline model.}
    \label{tab:hp}
\end{table}

\section{Original types to Common types}
\label{app:type_map}

\begin{table}[!ht]
    \centering
    \resizebox{0.5\linewidth}{!}{
        \begin{tabular}{cc}
            \toprule
                \textbf{Corpus-level} & \textbf{Common} \\
            \midrule
                PER & PER \\
                LOC & LOC \\
                FAC & FAC \\
                GPE & LOC \\
                VEH & Other \\
                ORG & ORG \\
            \bottomrule
            \end{tabular}
    }
    \caption{Litbank}
    \label{app:litbank}
\end{table}

\begin{table}[!ht]
    \centering
    \resizebox{0.5\linewidth}{!}{
        \begin{tabular}{cc}
            \toprule
                \textbf{Original} & \textbf{Common} \\
            \midrule
                PER & PER \\
                ORG & ORG \\
                LOC & LOC \\
                DIG & Other \\
            \bottomrule
            \end{tabular}
    }
    \caption{EmailCoref}
    \label{app:emailcoref}
\end{table}

\begin{table}[!ht]
    \centering
    \resizebox{0.5\linewidth}{!}{
        \begin{tabular}{cc}
            \toprule
                \textbf{Original} & \textbf{Common} \\
            \midrule
                Organization & ORG \\
                Person & PER \\
                Corporation & FAC \\
                Event & Other \\
                Place & LOC \\
                Thing & Other \\
                OTHER & Other \\
                NA & Other \\
            \bottomrule
            \end{tabular}
    }
    \caption{WikiCoref}
    \label{app:wikicoref}
\end{table}

\begin{table}[!ht]
    \centering
    \resizebox{0.5\linewidth}{!}{
        \begin{tabular}{cc}
            \toprule
                \textbf{Original} & \textbf{Common} \\
            \midrule
                ORG & ORG \\
                WORK\_OF\_ART & Other \\
                LOC & LOC \\
                CARDINAL & Other \\
                EVENT & Other \\
                NORP & Other \\
                GPE & LOC \\
                DATE & Other \\
                PERSON & PER \\
                FAC & FAC \\
                QUANTITY & Other \\
                ORDINAL & Other \\
                TIME & Other \\
                PRODUCT & Other \\
                PERCENT & Other \\
                MONEY & Other \\
                LAW & Other \\
                LANGUAGE & Other \\
                NA & Other \\
            \bottomrule
            \end{tabular}
    }
    \caption{OntoNotes}
    \label{app:ontonotes}
\end{table}

We use four common types (PERson, LOCation, FACility, ORGanization) and \textit{Other} in \textbf{+ET (com)} experiments. These types are annotated in most of the named-entity recognition datasets and therefore are easier to model and learn via machine learning approaches. Tables~\ref{app:litbank},~\ref{app:emailcoref},~\ref{app:wikicoref},~\ref{app:ontonotes} show the mapping between the original types of each coreference dataset used in our study to the reduced common types. The most drastic difference occurs for OntoNotes (19 -> 5) and WikiCoref (8 -> 5). \textit{OTHER} type in WikiCoref is for freebase links that did not have an associated type stored in freebase, whereas \textit{NA} is used for mentions which do not have a freebase link. For OntoNotes, \textit{NA} refers to the mentions that did not get any type assigned to them even after the use of our cluster based type-propagation approach (explained in Section~\ref{sec:data}).

\end{document}